\documentclass[11pt, a4paper, copyright]{weco_paper}

\usepackage{amsfonts}
\usepackage{amsmath}
\usepackage{wrapfig}
\usepackage{listings}
\usepackage{nicefrac}
\usepackage{float}
\usepackage{cleveref}
\usepackage{subcaption}
\usepackage{times}
\usepackage{latexsym}
\usepackage{xspace}
\usepackage{tabularx}
\makeatletter
\newcommand{\dashedmidrule}{%
  \noalign{\vskip\aboverulesep}%
  \noalign{\hbox to\textwidth{\leaders\hbox to 5pt{\hss\rule{3pt}{\lightrulewidth}\hss}\hfil}}%
  \noalign{\vskip\belowrulesep}}
\makeatother
\newcommand{\shadebox}[1]{{\setlength{\fboxsep}{1.1pt}\colorbox{gray!15}{#1}}}
\newcommand{\bpb}{\textsc{bpb}\xspace} 
\usepackage[T1]{fontenc}
\usepackage{booktabs}
\usepackage{array}
\usepackage{tabularx}
\usepackage{amsmath,amssymb}   % math
\usepackage{booktabs}          % \toprule \midrule \bottomrule
\usepackage{tabularx}          % the X column type (pulls in `array` for free)
\usepackage{hyperref}
\usepackage[utf8]{inputenc}
\usepackage{enumitem}
\usepackage{algorithm}
\usepackage{caption}
\usepackage{microtype}
\usepackage{makecell}
\usepackage{inconsolata}
\usepackage{caption}
\usepackage{multirow}
\usepackage{graphicx}
\usepackage[most]{tcolorbox}
\usepackage{listings}
\usepackage{xcolor}
\usepackage[most]{tcolorbox}
\usepackage{listings}
\usepackage[noend]{algpseudocode}

\algrenewcommand{\algorithmiccomment}[1]{\hfill\textit{\(\triangleright\)~#1}}
\newcommand{\algstage}[1]{\Statex\textit{\(\triangleright\)~#1}}

\lstdefinestyle{promptstyle}{
    basicstyle=\ttfamily\scriptsize,
    breaklines=true,
    breakatwhitespace=false,
    columns=fullflexible,
    keepspaces=true,
    frame=none
}

\newtcblisting{promptbox}[1][]{
    colback=gray!4,
    colframe=gray!45,
    arc=1mm,
    boxrule=0.4pt,
    left=1mm,
    right=1mm,
    top=1mm,
    bottom=1mm,
    listing only,
    listing options={style=promptstyle},
    title=#1
}
\definecolor{ideabg}{RGB}{229, 240, 252}     % light blue fill
\definecolor{ideaborder}{RGB}{120, 165, 220} % medium blue border
\definecolor{codebg}{RGB}{246, 250, 255}     % very pale blue for code

\lstdefinestyle{idealisting}{
    basicstyle=\ttfamily\footnotesize,
    backgroundcolor=\color{codebg},
    breaklines=true,
    showstringspaces=false,
    columns=fullflexible,
    keepspaces=true,
    language=Python,
    keywordstyle=\color{blue!70!black},
    commentstyle=\color{green!50!black}\itshape,
    stringstyle=\color{red!70!black},
    aboveskip=4pt,
    belowskip=4pt,
}

\newtcolorbox{ideabox}[1][]{
    colback=ideabg,
    colframe=ideaborder,
    boxrule=0.6pt,
    arc=2pt,
    left=8pt, right=8pt, top=6pt, bottom=6pt,
    fontupper=\small,
    #1
}
\title{\textit{AutoData}: Agentic Search for Pre-training Data Selection}
\author[1]{Yan Meng\textsuperscript{\dag}}
\author[2]{Dhruv Srikanth}
\author[2]{Bingchen Zhao}
\author[2]{Zhengyao Jiang}
\author[2]{Yuxiang Wu}

\affil[1]{University of Amsterdam}
\affil[2]{Weco AI}

\correspondingauthor{y.meng@uva.nl, yuxiang@weco.ai}

\begin{abstract}
LLM agents have recently shown promise in automating machine learning engineering by editing model and training code under execution feedback.
Data, however, remains largely outside this agentic optimisation loop.
We frame pre-training data selection as heuristic engineering over per-document features, i.e., lexical statistics, categorical labels, and perplexity. 
We introduce \textit{AutoData}, an agent that searches directly over executable selection algorithms.
Unlike prior data mixture methods that optimise weights over a fixed set of domains, \textit{AutoData} searches a richer program space of scoring, stratification, and stochastic selection rules, discovering feature interactions automatically by iteratively refining algorithms with validation feedback from a proxy model.
Within an overnight search, \textit{AutoData} discovers a selection algorithm that outperforms existing human-designed curation pipelines.
Despite being searched only on this small proxy, the discovered recipe transfers to larger scales and improves the downstream metric \textsc{CORE}.
These results suggest that data engineering can be treated as an agentic machine learning problem, extending autonomous research from model and training-code optimization to the data.

\end{abstract}

\begin{document}

\maketitle

\begingroup
    \renewcommand{\thefootnote}{\fnsymbol{footnote}}
    \footnotetext[2]{Work done during a research internship at Weco AI.}
\endgroup

\section{Introduction}
\label{sec:introduction}

The performance of large language models depends on both model architecture and data.
On the recent NanoChat leaderboard~\citep{karpathy2025nanochat}, training recipes have been discovered by auto-research agents \citep{karpathy2026autoresearch}, but these agents treat the training data as static. 
This overlooks a first-order lever, i.e., data. 
A recent NanoChat result shows that simply replacing the training data from FineWeb-Edu~\citep{penedo2024fineweb} with NVIDIA-ClimbMix~\citep{diao2025climb} reduces wall-clock GPT-2 training time by $27\%$, a larger gain than most architecture-level improvements at this scale.

In this paper, we close this gap by extending agentic search to data selection.
Existing human-designed data curation methods combine deduplication, quality classification, perplexity-based filtering, importance sampling, and domain mixture optimization~\citep{xie2023data, datacomp, penedo2024fineweb, ankner2025perplexed, thrush2025perplexity}. 
However, manually designed methods often rely on a fixed combination of these heuristics, thus bottlenecking both the design space and the iteration speed.

We introduce \textit{AutoData}, an agentic search framework for data selection.\footnote{Code will be released at \url{https://github.com/WecoAI/AutoData}}
We frame data selection as a heuristic engineering problem, where each document is characterized by a set of features, i.e., lexical statistics, categorical labels, perplexity, and LLM-annotated quality signals.
\textit{AutoData} builds on an AIDE-style~\citep{jiang2025aide}  LLM agent that searches over selection algorithms on these features.
At each step, the agent proposes an executable selection strategy, trains a small proxy model on the selected subset, observes validation feedback, and refines the strategy in the next iteration. 
\textit{AutoData} then returns the selection algorithm with the best proxy-validation performance.
%Since each proxy run takes roughly 10 minutes on a single H100 GPU, the overnight-loop economics drive auto-research to apply equally to data engineering.

Our results show that agentic search is an effective and practical paradigm for pre-training data engineering. 
Within $200$ search steps on a small GPT-2 proxy model ($125$M), \textit{AutoData} discovers recipes that outperform DCLM~\citep{datacomp}, perplexity filtering, RegMix~\citep{liu2025regmix}, and the default ClimbMix ordering \citep{diao2025climb} on both search objectives: including validation bits-per-byte (\textsc{val-bpb}) and downstream \textsc{CORE} accuracy. 
Without re-tuning, these discovered recipes transfer across model scales. \textit{AutoData} achieves the best \textsc{val-bpb} from $125$M to $897$M, with statistically significant improvement over all baselines. 
Analysis of the discovered recipes shows that they move beyond single-feature ranking and threshold filtering. 
Instead, \textit{AutoData} discovers composite scoring rules paired with diversity-preserving selection mechanisms, favoring higher-quality documents while maintaining broad coverage of the data distribution.

Overall, \textit{AutoData} shows that data engineering is a natural next frontier for autonomous AI research.
Instead of treating curation as a fixed, manually designed preprocessing pipeline, \textit{AutoData} makes data selection searchable: agents discover how to score documents, combine signals, and preserve diversity using direct validation feedback. 
This reframes pre-training data curation as an optimization problem over executable recipes, extending autonomous AI research to the data that shapes model learning.

%Third, cheap features alone already suffice: using only lexical, categorical, and perplexity features---none of which require training an additional model---\textit{AutoData} achieves measurable gains at small scale.
%Adding LLM-annotated features yields further improvement, suggesting that the paradigm scales with annotation richness and leaves substantial headroom for future work.

\begin{figure*}[!t]
    \centering
    \includegraphics[width=0.9\linewidth]{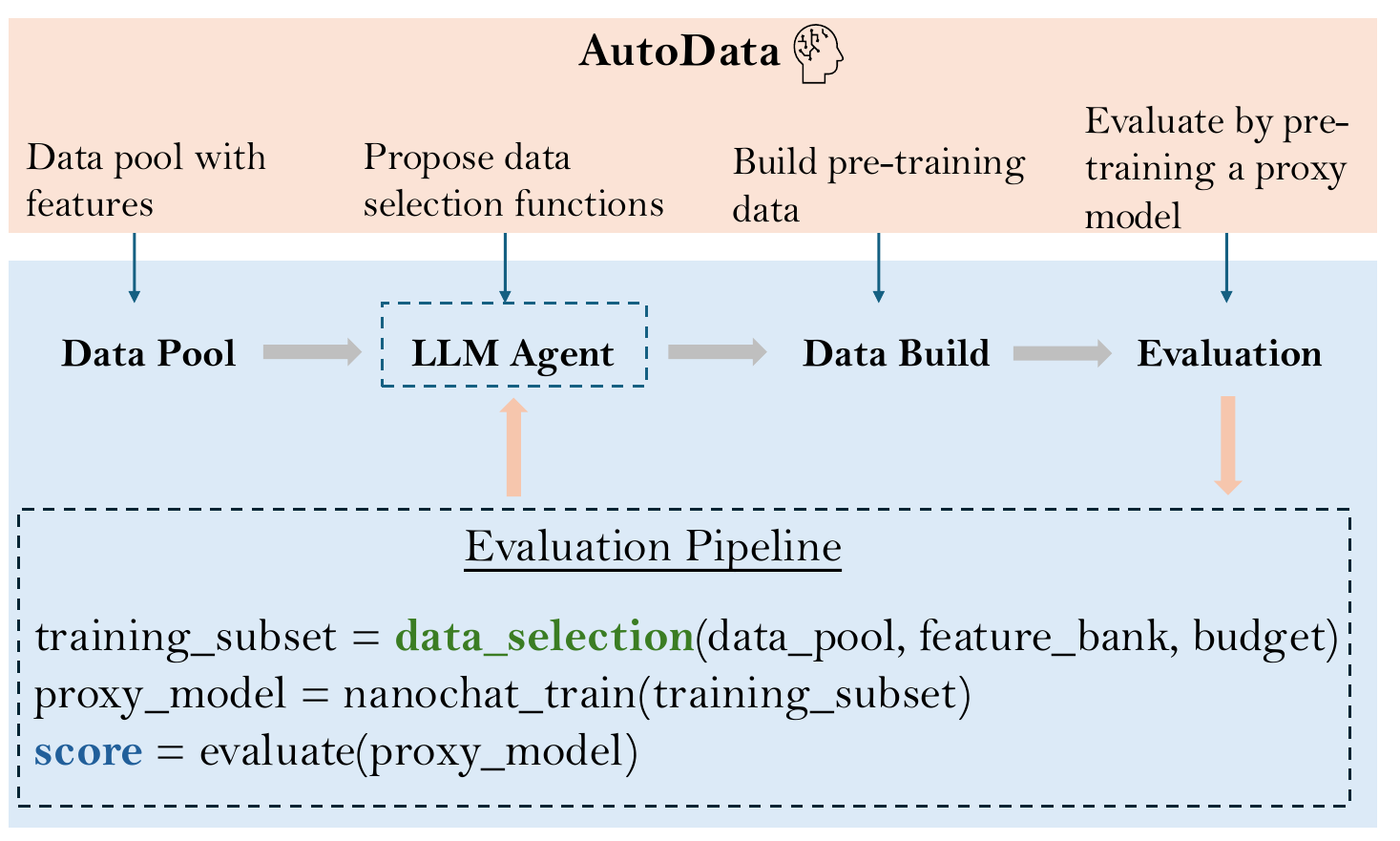}
  \caption{
Overview of \textit{AutoData}. An LLM agent proposes data-selection functions over a feature-annotated document pool. Each selected subset is used to pre-train a proxy language model, and the resulting validation score provides feedback for further search.
}
\label{fig:autodata-overview}
    \label{fig:placeholder}
\end{figure*}

%has shown that large web corpora benefit from human-designed curation pipelines based on deduplication, quality classification, perplexity-based pruning, importance sampling, and domain mixture optimisation \citep{xie2023data, datacomp, penedo2024fineweb, ankner2024perplexed, thrush2025perplexity}.
%These methods have produced high-quality pretraining datasets such as DCLM, FineWeb-Edu, and ClimbMix \citep{datacomp, penedo2024fineweb, diao2025climb}.

%In this work, we study data selection on top of such well-curated corpora.
%In this setting, most candidate documents are already plausible training examples.
%Aggressive filtering by a single quality signal can therefore remove useful diversity rather than simply discard noise.
%The objective is no longer only to identify low-quality documents, but to construct a training subset that provides the most useful learning signal under a fixed token budget.

\section{Background}

\subsection{NanoChat Task}

We use the NanoChat speedrun task as our experimental harness for language model pre-training~\citep{karpathy2025nanochat}.
NanoChat provides a lightweight GPT-style pre-training setup, and this small-scale setting makes iterative training experiments feasible. 
Moreover, this speedrun environment has produced insights that generalize beyond the leaderboard itself.
For example, Muon~\citep{jordan2024muon}, which originated from nanoGPT-style speedrun experiments, has recently challenged the dominance of AdamW~\citep{loshchilov2019decoupled} in large-scale language model training.

The leaderboard already shows that data quality is a first-order factor in training efficiency.
Simply switching the training corpus from FineWeb-Edu~\citep{penedo2024fineweb} to ClimbMix~\citep{diao2025climb} reduces the time required to reach the \textsc{CORE} threshold by roughly $27\%$.
However, the current speedrun uses only a small slice of ClimbMix---approximately the first $2\%$ of the full corpus.
This raises a natural question: under the same compute budget, whether we can select a more effective $2\%$ subset from the full ClimbMix pool.

\subsection{Base Corpus: NVIDIA ClimbMix}

We conduct our data selection experiments on ClimbMix~\citep{diao2025climb}, a 400B-token English pre-training corpus released by NVIDIA.
ClimbMix is produced by a human-designed pipeline, which is curated on Common Crawl using document clustering, reference-model scoring, and cluster-level reweighting.
We use the full ClimbMix as the source corpus and study whether LLM-generated selection strategies can identify $2\%$ compact subsets that outperform the default ones used in NanoChat.

\section{AutoData}

\textit{AutoData} builds on the AIDE scaffold~\citep{jiang2025aide}, an LLM-powered code-optimisation agent that improves candidate programs using execution feedback.
We adapt AIDE to the pre-training data selection task by optimising the selection algorithm via code. 
As shown in Figure~\ref{fig:autodata-overview}, the LLM agent proposes data selection methods with the given data pool, executes the selection function to build the training data, and evaluates the selection method by training a small proxy language model.
The evaluation metrics are provided to the LLM agent as feedback, guiding subsequent search toward more effective data selection methods.

\subsection{Problem Formulation}

We frame pre-training data selection as a search over selection algorithms. 
Given a candidate document pool $\mathcal{D}=\{d_i\}_{i=1}^{N}$ with document-level 
features $\{\mathbf{x}_i\}_{i=1}^{N}$ and a fixed budget $B$, a selection algorithm 
is a function
\[
f: \big(\mathcal{D},\, \{\mathbf{x}_i\}_{i=1}^{N},\, B\big) \rightarrow \mathcal{S}, 
\quad \mathcal{S} \subset \mathcal{D},\ |\mathcal{S}|=B,
\]
that maps the pool, feature bank, and budget to a budget-constrained subset. Let 
$\mathcal{A}_{\theta}(\mathcal{S})$ denote the model obtained by pre-training with 
fixed hyperparameters $\theta$ on $\mathcal{S}$, and let $J(\cdot)$ be an evaluation 
function (e.g., validation bits-per-byte or downstream task accuracy). \textit{AutoData} 
searches over the space $\mathcal{F}$ of selection algorithms to maximize empirical 
training performance:
\[
f^{*} = \arg\max_{f \in \mathcal{F}} \; J\!\left(\mathcal{A}_{\theta}\big(f(\mathcal{D}, 
\{\mathbf{x}_i\}, B)\big)\right).
\]
%Searching over algorithms rather than subsets directly grounds the search in a structured program space, where the agent composes multiple feature signals into interpretable selection rules.

\subsection{Feature Bank}

To support compositional selection, we provide the agent with document-level features along four complementary axes: lexical statistics, categorical labels, reference-model perplexity, and LLM-generated content annotations. 
These features characterize documents from multiple perspectives, capturing surface-level properties, topical diversity, learning difficulty as estimated by a reference model, and content factuality.

\paragraph{Lexical features.}
For each document, we compute its length in tokens (via the nanochat tokenizer) and distinct $n$-gram ratios for $n \in \{1, \ldots, 5\}$. Document length serves as a coarse quality signal: very short documents are often fragments, while very long documents tend to contain mixed or off-topic content. 
Distinct $n$-gram ratios capture repetition and templatic text. 
Low values tend to have boilerplate or repeated phrasing, while high values on short texts often flag broken or randomized strings.

\paragraph{Categorical features.}
We assign each document a topic label and a format label using the 
WebOrganizer classifiers.\footnote{\url{https://huggingface.co/WebOrganizer}} 
Topic and format together yield $24 \times 24 = 576$ joint categories on the ClimbMix pool, covering domains (e.g., {news, science, finance}) and document types (e.g., {article, FAQ, listicle, advertisement}).

\paragraph{Perplexity features.}
We compute per-document bits-per-byte under a small reference model 
(\texttt{Qwen2.5-0.5B-Base}). 
Motivated by prior work on perplexity-based pruning with small reference models~\citep{ankner2025perplexed}, this signal captures document difficulty relative to the reference distribution: low perplexity flags texts that are easy to predict while high perplexity flags noisy content.

\paragraph{LLM annotation features.}
For a subset of the candidate pool, we use LLM-generated annotations from \texttt{Gemini-3-Flash}.
These annotations extract three document-level counts:
\begin{itemize}[leftmargin=*, itemsep=2pt, topsep=2pt]
    \item \texttt{n\_factual}: number of incorrect factual claims;
    \item \texttt{n\_rsteps}: number of explicit inferential reasoning steps;
    \item \texttt{n\_rerrors}: number of invalid reasoning steps.
\end{itemize}

We aim to study whether content-level signals provide complementary information beyond cheap lexical, categorical, and perplexity features. 
Annotation guidelines are shown in Appendix \ref{sec:llm-annotate}.

\subsection{Self-Evolving Loop}

At each iteration, the LLM agent proposes a candidate selection function 
$f \in \mathcal{F}$, conditioned on a summary of previously evaluated 
candidates and their scores. 
The proposed program is executed in three stages: 
(i) it constructs a training subset 
$\mathcal{S} = f(\mathcal{D}, \{\mathbf{x}_i\}, B)$ from the document pool; 
(ii) the resulting subset is used to pre-train a proxy language model 
with fixed hyperparameters $\theta$; and 
(iii) the proxy is evaluated on a held-out validation set under an 
evaluation function $J(\cdot)$. 
The resulting score is returned to the agent as feedback, together with 
a summarized record of previous trajectories to inform the next proposal.

Through this self-evolving loop, \textit{AutoData} optimises pre-training 
data selection using empirical small-scale training performance as the 
objective signal. 
The specific proxy architecture, validation set, and 
evaluation metric used in our experiments are described in Section \ref{sec:experiments}.

\section{Experiments}
\label{sec:experiments}

\subsection{Experimental Setup}

\paragraph{Data pool.}
We apply \textit{AutoData} to NVIDIA's full ClimbMix corpus~\citep{diao2025climb}, a 400B-token English pretraining corpus.
The corpus contains $6{,}542$ training shards, covering $N=553{,}155{,}584$ documents.
We reserve one shard as a held-out validation set and exclude it from all selection pools.

\paragraph{Selection budget.}
\textit{AutoData} selects a subset of $B=14{,}374{,}266$ documents from the ClimbMix pool, corresponding to approximately $2.6\%$ of the corpus.
This budget matches the data scale used in the standard NanoChat pre-training task.
All methods are therefore compared under the same document-level selection budget.

\paragraph{Proxy model.}
Training a proxy model on the full $B$-document subset for every candidate strategy would be computationally expensive
We instead evaluate each strategy on an $880{,}000$-document subset sampled uniformly from its selected pool.
We train a depth-$8$ GPT-2 model with {target-param-data-ratio} $=10$, which corresponds to approximately $0.42$B training tokens.
Each run takes approximately $10$ minutes on one H100 GPU.
\begin{figure}[!t]
    \centering
    \begin{subfigure}[t]{0.49\linewidth}
        \centering
        \includegraphics[width=\linewidth]{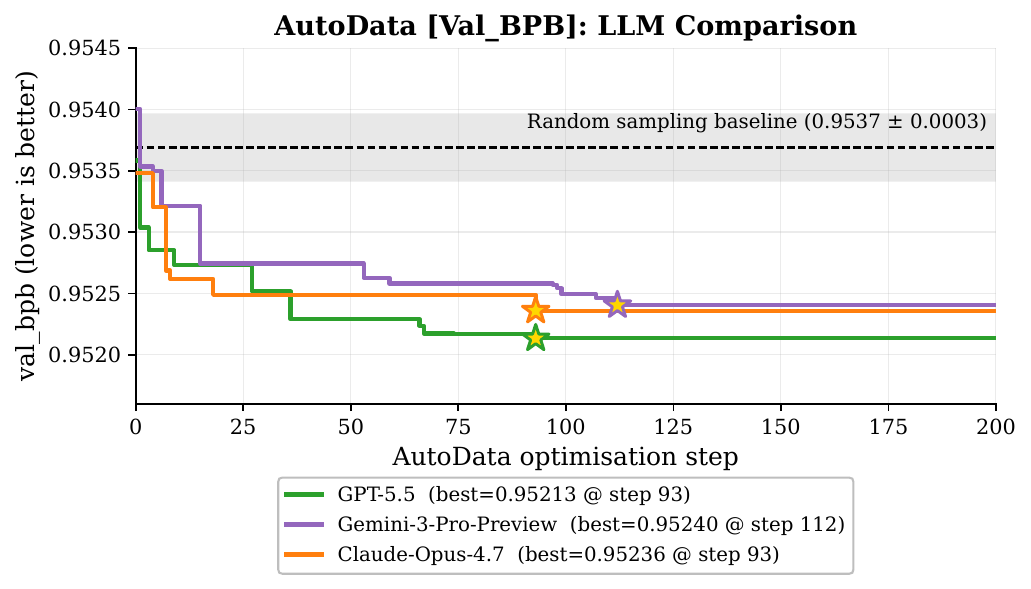}
        \caption{Optimisation signal on \textsc{val-bpb}.}
        \label{fig:base-llm-valbpb}
    \end{subfigure}\hfill
    \begin{subfigure}[t]{0.49\linewidth}
        \centering
        \includegraphics[width=\linewidth]{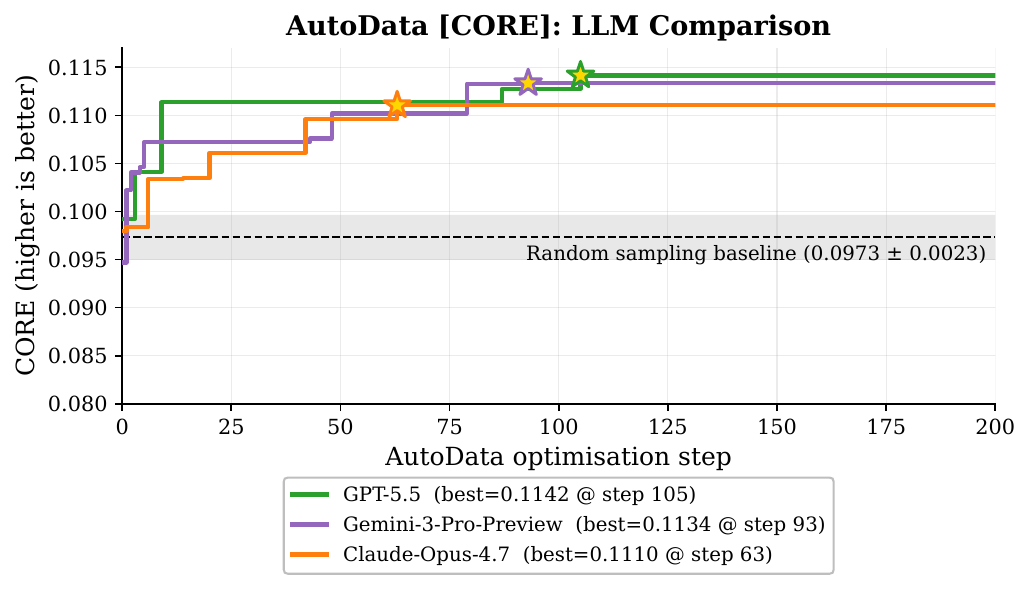}
        \caption{Optimisation signal on \textsc{CORE}.}
        \label{fig:base-llm-core}
    \end{subfigure}

    \caption{
        \textit{AutoData} search trajectories across base LLMs
        (GPT-5.5, Claude-Opus-4.7, and Gemini-3-Pro-Preview)
        on the full ClimbMix pool.
        Each step proposes a new data selection recipe, which is
        evaluated by training a proxy GPT-2 (depth=$8$) model.
    }
    \label{fig:base-llm-comparison}
\end{figure}

\paragraph{Subsample protocol.}
To reduce the high seed variance of single small-scale runs, we evaluate each selection strategy by training {four} proxy models on the same selected subsample, varying only the proxy training seed.
The four trainings run in parallel on $4{\times}$H100 GPUs.
Holding the subsample fixed and varying only the training seed isolates {training-induced variance}, yielding a stable estimate of each selector's performance.

\paragraph{Evaluation signals.}
\textit{AutoData} uses proxy-model performance as search feedback.
We consider two feedback signals: validation bits-per-byte (\textsc{val-bpb}) and downstream \textsc{CORE}.
For each signal, we run a separate \textit{AutoData} search using that signal as the optimization objective.
We then report both metrics for each discovered recipe: the optimized metric measures direct improvement, while the non-optimized metric tests whether the recipe generalizes beyond the feedback signal used during search.

\textsc{Val-bpb} is measured on a held-out ClimbMix shard, with lower values indicating better language modeling performance.
Following~\citet{datacomp}, \textsc{CORE} is used as the downstream evaluation metric.
It is defined as the unweighted mean of centered accuracy across 22 standard tasks spanning commonsense reasoning, reading comprehension, and world knowledge.

\subsection{Baselines}
\label{sec:baselines}

\paragraph{Random sampling.}
This baseline randomly samples documents from the ClimbMix selection pool.
It measures the performance of the curated ClimbMix corpus without applying data selection. 

\paragraph{DCLM classifier.}
We adapt the DCLM-Baseline classifier~\citep{datacomp} to the ClimbMix setting.
The original DCLM classifier is a FastText model trained with Reddit ELI5 and OpenHermes-2.5 as positive examples and RefinedWeb as negative examples.
We keep the positive data unchanged, but replace the negative data with documents sampled from ClimbMix so that the classifier is calibrated to our target corpus.
We then select documents based on the classifier score.

\paragraph{Perplexity filter.}
A perplexity-based filtering baseline using \texttt{Qwen2.5-0.5B} as 
the scoring model, following \citet{ankner2025perplexed}. 
Each document is scored by its perplexity under the reference model. 
Since ClimbMix is heavily pre-filtered, we follow the recommendation of 
\citet{ankner2025perplexed} for filtered corpora and select documents from the {middle} perplexity band. 

\paragraph{RegMix.}
We adapt RegMix~\citep{liu2025regmix} to ClimbMix using the $K{=}24$ WebOrganizer topic categories as domains.
We sample $N{=}128$ Dirichlet mixtures, train a GPT-2 (depth=$8$) proxy model on each, and fit a LightGBM regressor to predict \textsc{val-bpb} from mixture weights.
The mixture with the lowest predicted \textsc{val-bpb} is then used for topic sampling to form the final training sets.

\paragraph{Selection protocol.}
All methods follow a controlled two-stage protocol.
Each method first selects $B$ documents from the full ClimbMix pool, where $B$ is the global selection budget used by \textit{AutoData} and matches the training-data budget for GPT-2 depth-$24$.
For smaller models, including GPT-2 depth-$8$ and depth-$12$, we draw uniform subsamples from the same selected $B$-document subset according to their corresponding training budgets.
This keeps the source pool and global selection budget fixed across methods.
The training-data sizes for each model scale are reported in Appendix Table~\ref{tab:training-config}.

\subsection{AutoData Recipe}
\begin{algorithm}[!t]
\caption{\textsc{AutoData} (\textsc{val-bpb})}
\label{alg:autodata_valbpb}
\footnotesize
\begin{algorithmic}[1]
\Require pool $\mathcal{D}$ of size $N$, budget $B$, slate size $a=3$
\Ensure $B$ unique document indices
\State $B_{\text{back}} \gets \lfloor B/3 \rfloor$;\ \ $B_{\text{fill}} \gets B - B_{\text{back}}$
\State Draw $B_{\text{back}} + a B_{\text{fill}}$ docs from $\mathcal{D}$ w/o replacement
\State First $B_{\text{back}}$ form the \emph{backbone}; the rest form $B_{\text{fill}}$ slates of $a$
\algstage{Candidate scoring}
\State $s_{\text{div}} \gets \tfrac{1}{2}\!\left[\tanh\tilde{z}(\text{div}_{\text{avg}}) + \tanh\tilde{z}(\text{div}_{5})\right]$
\State $s_{\text{ppl}} \gets \tanh\tilde{z}(-\log \text{ppl})$
\State $s_{\text{len}} \gets -\lvert \tanh\tilde{z}(\log \text{tok}) \rvert$
\State $s_{\text{rat}} \gets -\lvert \tanh\tilde{z}(\text{chars}/\text{tok}) \rvert$
\State $s \gets s_{\text{div}} + s_{\text{ppl}}\tfrac{s_{\text{div}}+1}{2} + s_{\text{len}} + s_{\text{rat}}$
\algstage{Shrinkage centering, per topic$\times$format cell $c$}
\State $\mu_c \gets \operatorname{mean}(s \mid c)$;\ \ $\lambda_c \gets n_c/(n_c+\tilde{n})$
\State $s \gets s - \tfrac{1}{2}\lambda_c \mu_c$
\algstage{Gumbel tournament}
\State $s \gets s + g$, \quad $g \sim \text{Gumbel}(0,1)$
\State Take $\arg\max$ per slate; union with backbone
\State \Return sorted indices
\end{algorithmic}
\end{algorithm}

\textit{AutoData} generates $1200$ candidate selection methods across GPT-5.5, Gemini-3-Pro-Preview, and Claude-Opus-4.7.
For the main comparison against human-designed baselines (Section~\ref{sec:baselines}), we adopt the best recipe from each search objective (in Figure \ref{fig:base-llm-comparison}): the top recipe found when optimising \textsc{val-bpb} and \textsc{CORE}, both selected on the depth=$8$ proxy model.
The two recipes are shown in Algorithm~\ref{alg:autodata_valbpb} (\textsc{val-bpb}) and Algorithm~\ref{alg:autodata_core} (\textsc{CORE}).

\subsubsection{Algorithm 1 (\textsc{val-bpb})}

\paragraph{Stage 1: Scoring.}
One third of the budget is reserved as a uniformly sampled random backbone to ensure broad coverage.
For the remaining budget, each candidate receives a composite score combining four features: lexical diversity ($n$-gram), perplexity, document length, and character-to-token density.
A key design choice is that the perplexity term is gated by diversity: low perplexity weights higher only when $n$-gram diversity is also high, which prevents the score from favoring repetitive boilerplate.
Length and density functions as {central} terms that penalize outliers in either direction.

\paragraph{Stage 2: Selection.}
Scores are normalized within each (topic, format) cell via shrinkage-based centering, which reduces over-selection from frequent cells. 
Candidates are then partitioned into slates of three; Gumbel noise is added to the centered scores, and the best of each slate is selected, equivalent to softmax sampling.
The selected data balances topic coverage, quality-aware scoring, and stochastic competition.

\begin{table*}[!t]
  \centering
  \footnotesize
  \renewcommand{\arraystretch}{1.38}
  \setlength{\tabcolsep}{3.2pt}
  \begin{tabularx}{\textwidth}{@{} l *{5}{>{\centering\arraybackslash}X} @{}}
  \toprule
  \textbf{Method}
  & \textbf{d = $8$ ($125$M)}
  & \textbf{d = $12$ ($286$M)}
  & \textbf{d = $16$ ($537$M)}
  & \textbf{d = $20$ ($897$M)}
  & \textbf{d = $24$ ($1.3$B)}
  \\
  \midrule
  \multicolumn{6}{@{}l}{\textit{val-bpb} $\downarrow$} \\
  \dashedmidrule
  Random uniform
  & \shadebox{$0.9484_{\pm0.0002}$}
  & $0.8477_{\pm0.0003}$
  & \shadebox{$0.7806_{\pm0.0002}$}
  & \shadebox{$0.7346_{\pm0.0004}$}
  & $\mathbf{0.7055_{\pm0.0011}}$
  \\
  DCLM-Baseline
  & \shadebox{$0.9954_{\pm0.0001}$}
  & \shadebox{$0.8886_{\pm0.0001}$}
  & \shadebox{$0.8202_{\pm0.0003}$}
  & \shadebox{$0.7766_{\pm0.0006}$}
  & \shadebox{$0.7499_{\pm0.0001}$}
  \\
  PPL filter
  & \shadebox{$0.9605_{\pm0.0001}$}
  & \shadebox{$0.8638_{\pm0.0000}$}
  & \shadebox{$0.7983_{\pm0.0001}$}
  & \shadebox{$0.7545_{\pm0.0003}$}
  & \shadebox{$0.7267_{\pm0.0001}$}
  \\
  RegMix
  & \shadebox{$0.9550_{\pm0.0009}$}
  & \shadebox{$0.8529_{\pm0.0005}$}
  & \shadebox{$0.7848_{\pm0.0002}$}
  & \shadebox{$0.7401_{\pm0.0002}$}
  & \shadebox{$0.7124_{\pm0.0002}$}
  \\
  \midrule
  \textbf{AutoData (\textsc{CORE})}
  & $0.9529_{\pm0.0001}$
  & $0.8474_{\pm0.0001}$
  & $\mathbf{0.7791_{\pm0.0001}}$
  & $\mathbf{0.7334_{\pm0.0001}}$
  & $0.7064_{\pm0.0001}$
  \\
  \textbf{AutoData (\textsc{val-bpb})}
  & $\mathbf{0.9475_{\pm0.0000}}$
  & $\mathbf{0.8473_{\pm0.0001}}$
  & \shadebox{$0.7802_{\pm0.0001}$}
  & \shadebox{$0.7349_{\pm0.0004}$}
  & $0.7065_{\pm0.0003}$
  \\
  \midrule[1.2pt]
  \multicolumn{6}{@{}l}{\textit{CORE} $\uparrow$} \\
  \dashedmidrule
  
  Random uniform
  & $0.1041_{\pm0.0125}$
  & \shadebox{$0.1403_{\pm0.0034}$}
  & $\mathbf{0.2053_{\pm0.0065}}$
  & $0.2364_{\pm0.0011}$
  & $0.2609_{\pm0.0058}$
  \\
  DCLM-Baseline
  & $0.1043_{\pm0.0028}$
  & \shadebox{$0.1362_{\pm0.0065}$}
  & $0.2006_{\pm0.0111}$
  & \shadebox{$0.2245_{\pm0.0027}$}
  & $0.2470_{\pm0.0019}$
  \\
  PPL filter
  & \shadebox{$0.0933_{\pm0.0051}$}
  & \shadebox{$0.1449_{\pm0.0034}$}
  & $0.1926_{\pm0.0029}$
  & $0.2276_{\pm0.0080}$
  & $0.2535_{\pm0.0051}$
  \\
  RegMix
  & $0.0999_{\pm0.0080}$
  & $0.1508_{\pm0.0036}$
  & $0.2025_{\pm0.0105}$
  & $0.2319_{\pm0.0072}$
  & $0.2567_{\pm0.0081}$
  \\
  \midrule
  \textbf{AutoData (\textsc{CORE})}
  & $\mathbf{0.1142_{\pm0.0023}}$
  & $0.1430_{\pm0.0094}$
  & $0.2009_{\pm0.0015}$
  & $\mathbf{0.2389_{\pm0.0072}}$
  & $\mathbf{0.2727_{\pm0.0128}}$
  \\
  \textbf{AutoData (\textsc{val-bpb})}
  & $0.1037_{\pm0.0030}$
  & $\mathbf{0.1568_{\pm0.0021}}$
  & $0.1959_{\pm0.0004}$
  & $0.2372_{\pm0.0084}$
  & $0.2647_{\pm0.0085}$
  \\
  \bottomrule
  \end{tabularx}

  \caption{\textit{AutoData} v.s.\ human-designed data curation pipelines across five model scales (from depth = $8$ to $24$). We report the mean over three runs, with the subscript denoting the sample standard deviation. In each column, the best mean is shown in \textbf{bold}. A cell is \colorbox{gray!15}{shaded} when the
best \emph{tested} method is significantly better than it ($p < 0.05$). Tests are paired $t$-tests over the three training seeds.}
  \label{tab:cross-scale}
\end{table*}

\subsubsection{Algorithm 2 (\textsc{core})}

\paragraph{Stage 1: Scoring.}
The recipe first reserves $50\%$ of the budget as a uniformly sampled backbone to preserve corpus distribution.
For the remaining budget, it assigns each candidate document a composite score based on document length, character-to-token ratio, $n$-gram diversity, and perplexity.
These features are mapped to pool-adaptive scores: length and diversity prefer medium-to-high values, perplexity prefers fluent text, and the ratio score prefers normal text density.
The recipe further adds a prior by grouping candidates into contiguous source-order blocks and assigning each document the standardized mean composition score of its block.
The final score is the sum of the composition score, the prior, and Gumbel noise.

\paragraph{Stage 2: Selection.}
The final subset is the union of the random sampled backbone and repair set, followed by deduplication to obtain exactly $B$ unique documents.
The selected data balances data coverage, document-level quality, local source quality, and stochastic competition.

% =========================== Algorithm 2 ===================================
\begin{algorithm}[!t]
\caption{\textsc{AutoData} (\textsc{CORE})}
\label{alg:autodata_core}
\footnotesize
\begin{algorithmic}[1]
\Require pool $\mathcal{D}$ of size $N$, budget $B$, seed
\Ensure $B$ unique document indices
\State $B_{\text{back}} \gets \lfloor B/2 \rfloor$;\ \ $B_{\text{rep}} \gets B - B_{\text{back}}$
\State $n_{\text{cand}} \gets \min(N - B_{\text{back}},\, 2B_{\text{rep}})$
\State Draw $B_{\text{back}} + n_{\text{cand}}$ docs w/o replacement
\State First $B_{\text{back}}$ form the backbone; the rest form $\mathcal{C}$
\algstage{Finite-safe features}
\State $\ell_i \gets \log(1+\text{tok}_i)$;\ \ $r_i \gets \log\frac{\text{chars}_i+1}{\text{tok}_i+1}$
\State Load $\text{div}_i$, $\text{ppl}_i$ for $i \in \mathcal{C}$
\algstage{Pool-adaptive hygiene, $\operatorname{tri}$ = triangular kernel}
\State $s^{\text{len}}_i \gets \operatorname{tri}(\ell_i;\, q_{2/3}, q_{1/3}, q_{2/3})$
\State $s^{\text{div}}_i \gets \operatorname{tri}(\text{div}_i;\, q_{2/3}, q_{1/3}, q_{2/3})$
\State $s^{\text{ppl}}_i \gets \operatorname{tri}(\text{ppl}_i;\, q_{1/3}, q_{1/3}, q_{2/3})$
\State $s^{\text{rat}}_i \gets \operatorname{tri}(r_i;\, q_{1/2}, q_{1/3}, q_{2/3})$
\State $a_i \gets \operatorname{z}\!\big(\tfrac{1}{4}\textstyle\sum_{k} s^{k}_i\big)$
\algstage{Source-neighborhood prior}
\State $M \gets \max(1, \lfloor \sqrt{\lvert\mathcal{C}\rvert} \rfloor)$;\ \ $b_i \gets \min(\lfloor iM/N \rfloor, M{-}1)$
\State $\mu_b \gets \operatorname{mean}\{a_i : b_i = b\}$;\ \ $p_i \gets \operatorname{z}(\mu)_{b_i}$
\algstage{Stochastic repair}
\State $\text{score}_i \gets a_i + p_i + g_i$, \quad $g_i \sim \text{Gumbel}(0,1)$
\State $\mathcal{R} \gets \operatorname{Top}_{B_{\text{rep}}}(\mathcal{C};\, \text{score})$
\State $\mathcal{S} \gets \text{backbone} \cup \mathcal{R}$
\State \textbf{if} $\lvert\mathcal{S}\rvert \neq B$ \textbf{then} repair to exactly $B$ unique
\State \Return sorted indices $\mathcal{S}$
\end{algorithmic}
\end{algorithm}
\section{Results}

Tables~\ref{tab:cross-scale} compare \textit{AutoData} with human-designed data selection pipelines across five model scales for \textsc{val-bpb} and \textsc{CORE}. 
We highlight two findings.

\paragraph{Human-designed curation pipelines struggle on a well-curated pool.}
DCLM-Baseline and PPL filtering regress \textsc{val-bpb} across scales, indicating worse language modeling performance.
RegMix performs better on GPT-2 depth=$12$ \textsc{CORE}, but still does not consistently improve either \textsc{val-bpb} or \textsc{CORE}.
These results suggest that data curation methods are sensitive to the source data pool and may require redesign to remain effective when applied to already well-curated data.

\paragraph{{AutoData} transfers across model scales.}

\textit{AutoData} recipes significantly outperform all four baselines on \textsc{val-bpb} from depth $8$ to depth $20$, showing that recipes discovered with a small proxy model can remain effective at larger scales. 
At depth $24$, however, \textit{AutoData} performs similarly to the default ClimbMix, indicating a limit to its cross-scale transfer. 
On the downstream \textsc{CORE} benchmark, \textit{AutoData} outperforms the baselines at every scale except depth $16$. 
However, these gains are not statistically significant, possibly because \textsc{CORE} has high variance. 
The lack of statistically significant improvements on \textsc{CORE} is a limitation of our methodology.

\section{Analysis}

\subsection{Number of Feature Usage}
Table~\ref{tab:feature-usage} shows the number of features used in \textit{AutoData} generated recipes. 
First, all three LLMs use the same core features: length, $n$-gram, and perplexity.
Second, the LLMs disagree on categorical features (topic and format): GPT-5.5 and Opus-4.7 use them in about $90\%$ of recipes, while Gemini-3-Pro-Preview uses them in only $1$--$2\%$.
This reflects a real gap in how each agent thinks about the problem: GPT-5.5 and Opus-4.7 treat it as a compositional structure, while Gemini-3-Pro-Preview relies almost entirely on lexical statistics.

\begin{table}[!t]
\centering
\small
\setlength{\tabcolsep}{3.5pt}
\renewcommand{\arraystretch}{1.25}
\begin{tabular}{l ccc}
\toprule
\textbf{Feature} & \textbf{GPT-5.5} & \textbf{Gemini-3-Pro} & \textbf{Opus-4.7} \\
\midrule
Length     & $200$ & $200$ & $200$ \\
$N$-gram   & $200$ & $199$ & $200$ \\
Perplexity & $200$ & $199$ & $200$ \\
Topic      & $178$ & $2$ & $187$ \\
Format     & $177$ & $3$ & $172$ \\
\midrule
\textbf{Total recipes} & $200$ & $200$ & $200$ \\
\bottomrule
\end{tabular}
\caption{Number of feature usage in selection recipes generated by different LLMs. Here, we analyze recipes generated by \textit{AutoData} (\textsc{val-bpb}.)}
\label{tab:feature-usage}
\end{table}

\begin{table}[!t]
    \centering
    \small  % Changed from \scriptsize to \small
    \setlength{\tabcolsep}{5pt}  % Increased from 4pt
    \renewcommand{\arraystretch}{1.4}  % Increased from 1.15
    \begin{tabular}{@{}p{0.5\linewidth}rr@{}}  % Increased column width
    \toprule
    \multirow{2}{*}{\textbf{Feature subset}}
      & \multicolumn{2}{c}{\textbf{GPT-5.5}}  \\
    
      & \textbf{\bpb} $\downarrow$ & \textbf{$\Delta\sigma$} $\uparrow$
      \\
    \midrule
    Random sampling
      & \multicolumn{2}{c}{$0.95181 \pm 0.00077$} \\
    \midrule
    Lexical (2)
      & $0.95043$ & $+1.8$ \\
    Perplexity (1)
      & $0.95037$ & $+1.9$  \\
    Categorical (2)
      & $0.95041$ & $+1.8$ \\
    LLM annotations (3)
      & $\mathbf{0.95021}$ & $\mathbf{+2.1}$  \\
    All features (8)
      & $0.95048$ & $+1.7$  \\
    \bottomrule
    \end{tabular}
    \caption{Feature search-space ablation on the annotated subset data pool, where LLM-annotation features are available. \textbf{Bold} marks the best result.}
    \label{tab:feature-search-space-ablation}
\end{table}

\begin{table*}[!t]
\centering
\small
\setlength{\tabcolsep}{3.5pt}
\renewcommand{\arraystretch}{1.18}
\begin{tabularx}{\textwidth}{@{} l l X c c c @{}}
\toprule
\textbf{Axis} & \textbf{Sub-type} & \textbf{Operational definition}
& \textbf{GPT-5.5} & \textbf{Gemini-3-Pro} & \textbf{Opus-4.7} \\
\midrule
\multirow{4}{*}{\shortstack[l]{Score\\aggregation}}
& Single feature
& Use one normalized feature as the score: $s_i = x_{ik}$.
& $0.0\%$ & $0.0\%$ & $0.0\%$ \\
& Linear composite
& Sum normalized features with fixed weights:
  $s_i = \sum_k w_k\, x_{ik}$.
& $13.0\%$ & $\mathbf{96.0\%}$ & $\mathbf{85.5\%}$ \\
& Multiplicative composite
& Combine features by products or log-sums:
  $s_i = \prod_k \phi_k(x_{ik})^{w_k}$.
& $\mathbf{86.5\%}$ & $0.5\%$ & $1.0\%$ \\
& Gated / conditional
& Make a feature's weight depend on other features:
  $s_i = \sum_k w_k(x_i)\, x_{ik}$.
& $0.5\%$ & $3.5\%$ & $13.5\%$ \\
\midrule
\multirow{4}{*}{\shortstack[l]{Selection\\rule}}
& Threshold
& Select documents whose score passes a cutoff:
  $z_i = \mathbf{1}[s_i \geq \tau]$.
& $1.5\%$ & $6.0\%$ & $0.5\%$ \\
& Top-$B$
& Select the $B$ highest-scoring documents:
  $z_i = \mathbf{1}[i \in \operatorname{Top\text{-}B}(s)]$.
& $6.5\%$ & $6.5\%$ & $12.0\%$ \\
& Tournament
& Pick a winner per local slate $A$ by noisy comparison:
  $i^\star = \arg\max_{j \in A}(s_j+\epsilon_j)$.
& $\mathbf{91.0\%}$ & $0.0\%$ & $0.0\%$ \\
& Stratified quota
& Allocate a budget $B_c$ to each cell $c$, select within cells:
  $\sum_{i:\, c(i)=c} z_i = B_c$.
& $1.0\%$ & $\mathbf{87.5\%}$ & $\mathbf{87.5\%}$ \\
\bottomrule
\end{tabularx}
\caption{
Operation usage across LLM agents along two compositional axes, computed over $200$ recipes per LLM.
Within each axis, sub-types are mutually exclusive, so percentages sum to $100\%$ per agent (\textbf{bold} marks each LLM's dominant sub-type).
Notation: $x_i$ is the normalized feature vector of document $i$, $x_{ik}$ its $k$-th feature, $\phi_k$ a per-feature transform, $w_k$ a fixed weight, $s_i$ the document score, $z_i \in \{0,1\}$ the selection indicator, $\tau$ a threshold, $\epsilon_j$ noise, $c(i)$ the stratification cell of $i$, and $B_c$ the per-cell quota.
}
\label{tab:operation-usage}
\end{table*}

\subsection{Impact of Feature Space}

Table~\ref{tab:feature-search-space-ablation} reports the best \textsc{val-bpb} found by GPT-5.5 under five different feature subsets, evaluated against the random-sampling baseline.

\paragraph{LLM-annotation features yield the best overall result.} This suggests that semantic signals such as factual errors, reasoning errors, and reasoning steps provide useful structure beyond what cheap statistics can capture, and pointing to headroom for future work on richer annotations.

\paragraph{Cheap features alone already form a rich search space.}
Lexical, perplexity, and categorical subsets each individually reach $+1.8$--$1.9\sigma$ over random. 
For example, restricting \textit{AutoData} to only two lexical features ($n$-gram diversity and length) matches the improvement obtained from providing all eight features simultaneously, and a single perplexity feature does slightly better than the full set.
This indicates that \textit{AutoData} benefits from a small and well-chosen feature set. 
\subsection{Categories of \textit{AutoData} Recipes}
\label{sec:recipe-categories}
To interpret the methods generated by \textit{AutoData}, we categorize each recipe along two operational axes: {score aggregation} and {selection rule}. 
Let $\mathcal{D}$ be the candidate pool and $B$ the selection budget.
Each document $i \in \mathcal{D}$ has a normalized feature score $x_i = (x_{i1}, \ldots, x_{iK})$, where $x_{ik}$ is the feature value $k$.
A recipe first computes a score $s_i = g(x_i)$, then applies a selection rule to produce an indicator $z_i \in \{0,1\}$, with $z_i = 1$ if document $i$ is selected.

\paragraph{Annotation procedure.}
We annotate all $600$ recipes by \textit{AutoData} (\textsc{val-bpb}) ($200$ per LLM) along the two axes in Table~\ref{tab:operation-usage}. 
For each recipe, we provide Claude Opus-4.7 with the selection algorithm code and the operational definitions, and ask it to assign one mutually exclusive sub-type for score aggregation and one for the selection rule.

\paragraph{Score aggregation.}
The three LLM agents have different score aggregation types. 
GPT-5.5 prefers multiplicative composites ($86.5\%$), combining features through products or log-sum forms. 
Gemini-3-Pro-Preview and Opus-4.7 instead favor {linear composites} ($96.0\%$ and $85.5\%$), summing fixed-weight features into a single score.
Opus-4.7 uses gated scoring more than the others ($13.5\%$), suggesting a stronger tendency to vary the scoring rule across document groups.
No agent uses single-feature scoring, indicating that the recipes consistently treat selection as a multi-feature decision rather than filtering by one heuristic.

\paragraph{Selection rule.}
The LLMs also differ in selection rules.
GPT-5.5 primarily uses {tournament} selection ($91.0\%$), where each selection step compares a small candidate set and chooses the document with the highest noisy score. 
Unlike global top-$B$ selection, this rule favors high-scoring documents while still preserving stochasticity and diversity in the final subset.
Gemini-3-Pro-Preview and Opus-4.7 instead prefer {stratified quota} rules ($87.5\%$ each), partitioning the corpus into cells (by topic, format, or feature-value bins), allocating a quota $B_c$ per cell $c$, and selecting within each cell to preserve coverage.
Pure thresholding and global top-$B$ selection are rare across all LLMs, indicating they avoid hard filtering but in favor of rules that preserve diversity.

\section{Related Work}

\paragraph{Pre-training data curation and configuration search.}
Prior works have developed pipelines for curating high-quality data. 
Early large-scale corpora such as C4 \citep{raffel2020exploring} relied on rule-based cleaning, language identification, etc. 
More recent open corpora, including Dolma, DataComp-LM, FineWeb, FineWeb-Edu, and ClimbMix, make these design choices more explicit and evaluate how filtering, deduplication, and source composition affect downstream model quality~\citep{soldaini-etal-2024-dolma,datacomp,penedo2024fineweb,diao2025climb}.
These pipelines combine heuristics such as deduplication~\citep{lee-etal-2022-deduplicating,Tirumala2023D4IL}, quality filtering~\citep{datacomp,penedo2024fineweb}, classifier- or model-based filtering~\citep{datacomp}, loss-based data selection~\citep{ankner2025perplexed,thrush2025perplexity}, training data correction~\citep{meng-etal-2025-learn}, importance resampling toward a target distribution~\citep{xie2023data}, and domain-mixture optimisation~\citep{xie2023doremi,liu2025regmix,diao2025climb,ye2025data}.
Another line of work studies whether small-scale proxy experiments can predict better data choices at a larger scale~\citep{liu2025regmix,magnusson2025datadecide}.

\paragraph{LLM agents for research automation.}
A growing line of work uses LLM agents to automate machine learning and scientific research.
AIDE~\citep{jiang2025aide} frames ML engineering as a tree search over executable code, where an LLM proposes, runs, and refines candidate solutions using validation feedback.
Related systems automate ML experimentation and data-science workflows~\citep{huang2024mlagentbench, chan2024mlebench, guo2024dsagent, hong-etal-2025-data}, research ideation~\citep{baek-etal-2025-researchagent}, autonomous ML research workflows~\citep{li2024mlrcopilot, schmidgall-etal-2025-agent}, and automated scientific discovery or paper generation~\citep{lu2024aiscientist, Yamada2025TheAS}.
Recent work also studies research agents as search policies over candidate ML solutions~\citep{baek-etal-2025-researchagent}.
These systems mainly automate model development, experiment design, research ideation, or end-to-end research workflows.
\textit{AutoData} applies the same propose--execute--refine paradigm to pre-training data engineering, where each candidate solution is an executable data-selection method.

\section{Conclusion}

We introduce \textit{AutoData}, an agentic framework for searching pre-training data selection algorithms.
This work extends autonomous research from model-side optimisation to data engineering, studying whether LLM agents can discover effective data engineering strategies.
Built on an AIDE-style LLM agent, \textit{AutoData} treats each candidate as an executable selection recipe over document-level features and refines these recipes through proxy training feedback.
With a few hundred search steps on a small GPT-$2$ proxy model, \textit{AutoData} discovers recipes that outperform human-designed data curation baselines.
The discovered recipes further transfer across several model scales and evaluation metrics, suggesting that small-scale proxy search can produce useful data selection strategies for larger pre-training runs.
Overall, our results position pre-training data selection as a promising target for auto-research and show that LLM-generated recipes can serve as effective alternatives to human-designed curation pipelines.

\section*{Limitations}

We acknowledge several limitations of this work.
First, our experiments are conducted on models from $125$M to $1.3$B parameters and use a single base corpus, ClimbMix.
Although the discovered recipes transfer across model scales within this setting, their generalization to other corpora, domains, and larger model scales remains an open question.
Second, the effectiveness of \textit{AutoData} depends on the proxy objective used during search. 
Using \textsc{CORE} and \textsc{val-bpb} as search objectives yields different levels of improvement, highlighting the importance of choosing reliable proxy objectives for agentic data selection.
Finally, some of the strongest gains rely on LLM-annotated document features, whose annotation cost is non-trivial.
Understanding the cost--quality trade-off across feature sources, and scaling \textit{AutoData} to richer annotations and larger data pools, are important directions for future work.

%On the NanoChat benchmark, \textit{AutoData} finds effective recipes within $200$ search steps on a $125$M proxy model. 
%These recipes transfer without re-tuning to models up to $1.3$B parameters, achieving strong validation loss and improving downstream \textsc{CORE} by up to $+1.65\%$ on held-out tasks that are never used during search. 
%The discovered recipes consistently move beyond single-feature filtering, instead using composite scoring and diversity-preserving selection rules. 

\section*{Acknowledgments}
This research was supported by the NVIDIA DGX Cloud Innovation Lab. We also thank the reviewers for their valuable feedback and suggestions.

{\small
\bibliography{paper}

@inproceedings{
penedo2024fineweb,
title={The FineWeb Datasets: Decanting the Web for the Finest Text Data at Scale},
author={Guilherme Penedo and Hynek Kydl{\'\i}{\v{c}}ek and Loubna Ben allal and Anton Lozhkov and Margaret Mitchell and Colin Raffel and Leandro Von Werra and Thomas Wolf},
booktitle={The Thirty-eight Conference on Neural Information Processing Systems Datasets and Benchmarks Track},
year={2024},
url={https://openreview.net/forum?id=n6SCkn2QaG}
}

@inproceedings{xie2023data,
  title        = {Data Selection for Language Models via Importance Resampling},
  author       = {Xie, Sang Michael and Santurkar, Shibani and Ma, Tengyu and Liang, Percy},
  booktitle    = {Advances in Neural Information Processing Systems},
  year         = {2023},
  url          = {https://openreview.net/forum?id=uPSQv0leAu}
}

@inproceedings{
xie2023doremi,
title={DoReMi: Optimizing Data Mixtures Speeds Up Language Model Pretraining},
author={Sang Michael Xie and Hieu Pham and Xuanyi Dong and Nan Du and Hanxiao Liu and Yifeng Lu and Percy Liang and Quoc V Le and Tengyu Ma and Adams Wei Yu},
booktitle={Thirty-seventh Conference on Neural Information Processing Systems},
year={2023},
url={https://openreview.net/forum?id=lXuByUeHhd}
}

@inproceedings{
ankner2025perplexed,
title={Perplexed by Perplexity: Perplexity-Based Data Pruning With Small Reference Models},
author={Zachary Ankner and Cody Blakeney and Kartik Sreenivasan and Max Marion and Matthew L Leavitt and Mansheej Paul},
booktitle={The Thirteenth International Conference on Learning Representations},
year={2025},
url={https://openreview.net/forum?id=1GTARJhxtq}
}

@inproceedings{
thrush2025perplexity,
title={Improving Pretraining Data Using Perplexity Correlations},
author={Tristan Thrush and Christopher Potts and Tatsunori Hashimoto},
booktitle={The Thirteenth International Conference on Learning Representations},
year={2025},
url={https://openreview.net/forum?id=huuKoVQnB0}
}

@article{jiang2025aide,
      title={AIDE: AI-Driven Exploration in the Space of Code}, 
      author={Zhengyao Jiang and Dominik Schmidt and Dhruv Srikanth and Dixing Xu and Ian Kaplan and Deniss Jacenko and Yuxiang Wu},
      year={2025},
      eprint={2502.13138},
      archivePrefix={arXiv},
      primaryClass={cs.AI},
      url={https://arxiv.org/abs/2502.13138}, 
}

@misc{karpathy2025nanochat,
  author       = {Karpathy, Andrej},
  title        = {nanochat: The best ChatGPT that \$100 can buy},
  year         = {2025},
  howpublished = {\url{https://github.com/karpathy/nanochat}},
  note         = {GitHub repository}
}

@misc{karpathy2026autoresearch,
  author       = {Karpathy, Andrej},
  title        = {autoresearch},
  year         = {2026},
  howpublished = {\url{https://github.com/karpathy/autoresearch}},
  note         = {GitHub repository}
}

@inproceedings{lee-etal-2022-deduplicating,
    title = "Deduplicating Training Data Makes Language Models Better",
    author = "Lee, Katherine  and
      Ippolito, Daphne  and
      Nystrom, Andrew  and
      Zhang, Chiyuan  and
      Eck, Douglas  and
      Callison-Burch, Chris  and
      Carlini, Nicholas",
    editor = "Muresan, Smaranda  and
      Nakov, Preslav  and
      Villavicencio, Aline",
    booktitle = "Proceedings of the 60th Annual Meeting of the Association for Computational Linguistics (Volume 1: Long Papers)",
    month = may,
    year = "2022",
    address = "Dublin, Ireland",
    publisher = "Association for Computational Linguistics",
    url = "https://aclanthology.org/2022.acl-long.577/",
    doi = "10.18653/v1/2022.acl-long.577",
    pages = "8424--8445"
}

@article{Tirumala2023D4IL,
  title={D4: Improving LLM Pretraining via Document De-Duplication and Diversification},
  author={Kushal Tirumala and Daniel Simig and Armen Aghajanyan and Ari S. Morcos},
  journal={ArXiv},
  year={2023},
  volume={abs/2308.12284},
  url={https://api.semanticscholar.org/CorpusID:261076313}
}

@article{li2024mlrcopilot,
  title={MLR-Copilot: Autonomous Machine Learning Research based on Large Language Models Agents},
  author={Ruochen Li and Teerth Patel and Qingyun Wang and Qingyun Wang and Xinya Du},
  journal={ArXiv},
  year={2024},
  volume={abs/2408.14033},
  url={https://api.semanticscholar.org/CorpusID:271957477}
}

@article{lu2024aiscientist,
  title={The AI Scientist: Towards Fully Automated Open-Ended Scientific Discovery},
  author={Chris Lu and Cong Lu and Robert Tjarko Lange and Jakob N. Foerster and Jeff Clune and David Ha},
  journal={ArXiv},
  year={2024},
  volume={abs/2408.06292},
  url={https://api.semanticscholar.org/CorpusID:271854887}
}

@misc{jordan2024muon,
  title        = {Muon: An optimizer for hidden layers in neural networks},
  author       = {Jordan, Keller and Jin, Yuchen and Boza, Vlado and You, Jiacheng and Cesista, Franz and Newhouse, Laker and Bernstein, Jeremy},
  year         = {2024},
  howpublished = {\url{https://kellerjordan.github.io/posts/muon/}},
  note         = {Blog post, accessed 2026-05-22}
}

@inproceedings{loshchilov2019decoupled,
  title     = {Decoupled Weight Decay Regularization},
  author    = {Loshchilov, Ilya and Hutter, Frank},
  booktitle = {International Conference on Learning Representations},
  year      = {2019},
  url       = {https://openreview.net/forum?id=Bkg6RiCqY7}
}

@inproceedings{
huang2024mlagentbench,
title={{MLA}gentBench: Evaluating Language Agents on Machine Learning Experimentation},
author={Qian Huang and Jian Vora and Percy Liang and Jure Leskovec},
booktitle={Forty-first International Conference on Machine Learning},
year={2024},
url={https://openreview.net/forum?id=1Fs1LvjYQW}
}

@inproceedings{
guo2024dsagent,
title={{DS}-Agent: Automated Data Science by Empowering Large Language Models with Case-Based Reasoning},
author={Siyuan Guo and Cheng Deng and Ying Wen and Hechang Chen and Yi Chang and Jun Wang},
booktitle={Forty-first International Conference on Machine Learning},
year={2024},
url={https://openreview.net/forum?id=LfJgeBNCFI}
}

@inproceedings{hong-etal-2025-data,
    title = "Data Interpreter: An {LLM} Agent for Data Science",
    author = "Hong, Sirui  and
      Lin, Yizhang  and
      Liu, Bang  and
      Liu, Bangbang  and
      Wu, Binhao  and
      Zhang, Ceyao  and
      Li, Danyang  and
      Chen, Jiaqi  and
      Zhang, Jiayi  and
      Wang, Jinlin  and
      Zhang, Li  and
      Zhang, Lingyao  and
      Yang, Min  and
      Zhuge, Mingchen  and
      Guo, Taicheng  and
      Zhou, Tuo  and
      Tao, Wei  and
      Tang, Robert  and
      Lu, Xiangtao  and
      Zheng, Xiawu  and
      Liang, Xinbing  and
      Fei, Yaying  and
      Cheng, Yuheng  and
      Ni, Yongxin  and
      Gou, Zhibin  and
      Xu, Zongze  and
      Luo, Yuyu  and
      Wu, Chenglin",
    editor = "Che, Wanxiang  and
      Nabende, Joyce  and
      Shutova, Ekaterina  and
      Pilehvar, Mohammad Taher",
    booktitle = "Findings of the Association for Computational Linguistics: ACL 2025",
    month = jul,
    year = "2025",
    address = "Vienna, Austria",
    publisher = "Association for Computational Linguistics",
    url = "https://aclanthology.org/2025.findings-acl.1016/",
    doi = "10.18653/v1/2025.findings-acl.1016",
    pages = "19796--19821",
    ISBN = "979-8-89176-256-5"
}

@inproceedings{schmidgall-etal-2025-agent,
    title = "Agent Laboratory: Using {LLM} Agents as Research Assistants",
    author = "Schmidgall, Samuel  and
      Su, Yusheng  and
      Wang, Ze  and
      Sun, Ximeng  and
      Wu, Jialian  and
      Yu, Xiaodong  and
      Liu, Jiang  and
      Moor, Michael  and
      Liu, Zicheng  and
      Barsoum, Emad",
    editor = "Christodoulopoulos, Christos  and
      Chakraborty, Tanmoy  and
      Rose, Carolyn  and
      Peng, Violet",
    booktitle = "Findings of the Association for Computational Linguistics: EMNLP 2025",
    month = nov,
    year = "2025",
    address = "Suzhou, China",
    publisher = "Association for Computational Linguistics",
    url = "https://aclanthology.org/2025.findings-emnlp.320/",
    doi = "10.18653/v1/2025.findings-emnlp.320",
    pages = "5977--6043",
    ISBN = "979-8-89176-335-7"
}

@article{Yamada2025TheAS,
  title={The AI Scientist-v2: Workshop-Level Automated Scientific Discovery via Agentic Tree Search},
  author={Yutaro Yamada and Robert Tjarko Lange and Cong Lu and Shengran Hu and Chris Lu and Jakob N. Foerster and Jeff Clune and David Ha},
  journal={ArXiv},
  year={2025},
  volume={abs/2504.08066},
  url={https://api.semanticscholar.org/CorpusID:277741107}
}

@article{raffel2020exploring,
author = {Raffel, Colin and Shazeer, Noam and Roberts, Adam and Lee, Katherine and Narang, Sharan and Matena, Michael and Zhou, Yanqi and Li, Wei and Liu, Peter J.},
title = {Exploring the limits of transfer learning with a unified text-to-text transformer},
year = {2020},
issue_date = {January 2020},
publisher = {JMLR.org},
volume = {21},
number = {1},
issn = {1532-4435},
journal = {J. Mach. Learn. Res.},
month = jan,
articleno = {140},
numpages = {67}
}

@inproceedings{soldaini-etal-2024-dolma,
    title = "Dolma: an Open Corpus of Three Trillion Tokens for Language Model Pretraining Research",
    author = "Soldaini, Luca  and
      Kinney, Rodney  and
      Bhagia, Akshita  and
      Schwenk, Dustin  and
      Atkinson, David  and
      Authur, Russell  and
      Bogin, Ben  and
      Chandu, Khyathi  and
      Dumas, Jennifer  and
      Elazar, Yanai  and
      Hofmann, Valentin  and
      Jha, Ananya  and
      Kumar, Sachin  and
      Lucy, Li  and
      Lyu, Xinxi  and
      Lambert, Nathan  and
      Magnusson, Ian  and
      Morrison, Jacob  and
      Muennighoff, Niklas  and
      Naik, Aakanksha  and
      Nam, Crystal  and
      Peters, Matthew  and
      Ravichander, Abhilasha  and
      Richardson, Kyle  and
      Shen, Zejiang  and
      Strubell, Emma  and
      Subramani, Nishant  and
      Tafjord, Oyvind  and
      Walsh, Evan  and
      Zettlemoyer, Luke  and
      Smith, Noah  and
      Hajishirzi, Hannaneh  and
      Beltagy, Iz  and
      Groeneveld, Dirk  and
      Dodge, Jesse  and
      Lo, Kyle",
    editor = "Ku, Lun-Wei  and
      Martins, Andre  and
      Srikumar, Vivek",
    booktitle = "Proceedings of the 62nd Annual Meeting of the Association for Computational Linguistics (Volume 1: Long Papers)",
    month = aug,
    year = "2024",
    address = "Bangkok, Thailand",
    publisher = "Association for Computational Linguistics",
    url = "https://aclanthology.org/2024.acl-long.840/",
    doi = "10.18653/v1/2024.acl-long.840",
    pages = "15725--15788"
}

@inproceedings{
ye2025data,
title={Data Mixing Laws: Optimizing Data Mixtures by Predicting Language Modeling Performance},
author={Jiasheng Ye and Peiju Liu and Tianxiang Sun and Jun Zhan and Yunhua Zhou and Xipeng Qiu},
booktitle={The Thirteenth International Conference on Learning Representations},
year={2025},
url={https://openreview.net/forum?id=jjCB27TMK3}
}

@inproceedings{
magnusson2025datadecide,
title={DataDecide: How to Predict Best Pretraining Data with Small Experiments},
author={Ian Magnusson and Nguyen Tai and Ben Bogin and David Heineman and Jena D. Hwang and Luca Soldaini and Akshita Bhagia and Jiacheng Liu and Dirk Groeneveld and Oyvind Tafjord and Noah A. Smith and Pang Wei Koh and Jesse Dodge},
booktitle={Forty-second International Conference on Machine Learning},
year={2025},
url={https://openreview.net/forum?id=p9YlQPF8fE}
}

@inproceedings{meng-etal-2025-learn,
    title = "How to Learn in a Noisy World? Self-Correcting the Real-World Data Noise in Machine Translation",
    author = "Meng, Yan  and
      Wu, Di  and
      Monz, Christof",
    editor = "Chiruzzo, Luis  and
      Ritter, Alan  and
      Wang, Lu",
    booktitle = "Findings of the Association for Computational Linguistics: NAACL 2025",
    month = apr,
    year = "2025",
    address = "Albuquerque, New Mexico",
    publisher = "Association for Computational Linguistics",
    url = "https://aclanthology.org/2025.findings-naacl.416/",
    doi = "10.18653/v1/2025.findings-naacl.416",
    pages = "7466--7482",
    ISBN = "979-8-89176-195-7"
}

@inproceedings{
liu2025regmix,
title={RegMix: Data Mixture as Regression for Language Model Pre-training},
author={Qian Liu and Xiaosen Zheng and Niklas Muennighoff and Guangtao Zeng and Longxu Dou and Tianyu Pang and Jing Jiang and Min Lin},
booktitle={The Thirteenth International Conference on Learning Representations},
year={2025},
url={https://openreview.net/forum?id=5BjQOUXq7i}
}

@inproceedings{datacomp,
 author = {Li, Jeffrey and Fang, Alex and Smyrnis, Georgios and Ivgi, Maor and Jordan, Matt and Gadre, Samir and Bansal, Hritik and Guha, Etash and Keh, Sedrick and Arora, Kushal and Garg, Saurabh and Xin, Rui and Muennighoff, Niklas and Heckel, Reinhard and Mercat, Jean and Chen, Mayee and Gururangan, Suchin and Wortsman, Mitchell and Albalak, Alon and Bitton, Yonatan and Nezhurina, Marianna and Abbas, Amro and Hsieh, Cheng-Yu and Ghosh, Dhruba and Gardner, Josh and Kilian, Maciej and Zhang, Hanlin and Shao, Rulin and Pratt, Sarah and Sanyal, Sunny and Ilharco, Gabriel and Daras, Giannis and Marathe, Kalyani and Gokaslan, Aaron and Zhang, Jieyu and Chandu, Khyathi and Nguyen, Thao and Vasiljevic, Igor and Kakade, Sham and Song, Shuran and Sanghavi, Sujay and Faghri, Fartash and Oh, Sewoong and Zettlemoyer, Luke and Lo, Kyle and El-Nouby, Alaaeldin and Pouransari, Hadi and Toshev, Alexander and Wang, Stephanie and Groeneveld, Dirk and Soldaini, Luca and Koh, Pang Wei and Jitsev, Jenia and Kollar, Thomas and Dimakis, Alexandros G. and Carmon, Yair and Dave, Achal and Schmidt, Ludwig and Shankar, Vaishaal},
 booktitle = {Advances in Neural Information Processing Systems},
 doi = {10.52202/079017-0455},
 editor = {A. Globerson and L. Mackey and D. Belgrave and A. Fan and U. Paquet and J. Tomczak and C. Zhang},
 pages = {14200--14282},
 publisher = {Curran Associates, Inc.},
 title = {DataComp-LM: In search of the next generation of training sets for language models},
 url = {https://proceedings.neurips.cc/paper_files/paper/2024/file/19e4ea30dded58259665db375885e412-Paper-Datasets_and_Benchmarks_Track.pdf},
 volume = {37},
 year = {2024}
}

@inproceedings{diao2025climb,
  archiveprefix = {arxiv},
  eprint        = {2504.13161},
    title={CLIMB: Clustering-based Iterative Data Mixture Bootstrapping for Language Model Pre-training},
    author={Shizhe Diao and Yu Yang and Yonggan Fu and Xin Dong and Dan Su and Markus Kliegl and Zijia Chen and Peter Belcak and Yoshi Suhara and Hongxu Yin and Mostofa Patwary and Yingyan Celine Lin and Jan Kautz and Pavlo Molchanov},
  booktitle  = {Advances in Neural Information Processing Systems (NeurIPS)},
  month      = {December},
  year       = {2025},
}

@inproceedings{
chan2024mlebench,
title={{MLE}-bench: Evaluating Machine Learning Agents on Machine Learning Engineering},
author={Jun Shern Chan and Neil Chowdhury and Oliver Jaffe and James Aung and Dane Sherburn and Evan Mays and Giulio Starace and Kevin Liu and Leon Maksin and Tejal Patwardhan and Aleksander Madry and Lilian Weng},
booktitle={The Thirteenth International Conference on Learning Representations},
year={2025},
url={https://openreview.net/forum?id=6s5uXNWGIh}
}

@inproceedings{baek-etal-2025-researchagent,
    title = "{R}esearch{A}gent: Iterative Research Idea Generation over Scientific Literature with Large Language Models",
    author = "Baek, Jinheon  and
      Jauhar, Sujay Kumar  and
      Cucerzan, Silviu  and
      Hwang, Sung Ju",
    editor = "Chiruzzo, Luis  and
      Ritter, Alan  and
      Wang, Lu",
    booktitle = "Proceedings of the 2025 Conference of the Nations of the Americas Chapter of the Association for Computational Linguistics: Human Language Technologies (Volume 1: Long Papers)",
    month = apr,
    year = "2025",
    address = "Albuquerque, New Mexico",
    publisher = "Association for Computational Linguistics",
    url = "https://aclanthology.org/2025.naacl-long.342/",
    doi = "10.18653/v1/2025.naacl-long.342",
    pages = "6709--6738",
    ISBN = "979-8-89176-189-6"
}
}

\appendix

\section{Appendix}
\label{sec:appendix}

\subsection{The use of AI Assistant}
We used AI assistants for auxiliary support during the preparation of this paper. Specifically, we used them to improve grammar, clarity, and wording in the manuscript, and to help debug code used in our experiments. All research ideas, experimental designs, analyses, and conclusions were developed and verified by the authors.

\subsection{Training Details}

Table~\ref{tab:training-config} summarizes the model sizes and training budgets used in our cross-scale experiments.

\begin{table*}[!ht]
  \centering
  \small
  \setlength{\tabcolsep}{6pt}
  \renewcommand{\arraystretch}{1.15}
  \begin{tabular}{@{}l c c c c c@{}}
  \toprule
  \textbf{Depth}
  & \textbf{Total params}
  & \textbf{Scaling params\textsuperscript{\dag}}
  & \textbf{Ratio $r$}
  & \textbf{Training tokens} \\
  \midrule
  d8   & $125.8$\,M & $41.9$\,M  & $10$ & $0.42$\,B \\
  d12  & $286.3$\,M & $110.1$\,M & $10$ & $1.10$\,B \\
  d16  & $536.9$\,M & $234.9$\,M & $10$ & $2.35$\,B  \\
  d20  & $896.5$\,M & $435.2$\,M & $10$ & $4.35$\,B  \\
  d24  & $1.38$\,B  & $729.8$\,M & $8$  & $5.84$\,B \\
  \bottomrule
  \end{tabular}
  \caption{
  Training configurations across the five model scales.
  \textsuperscript{\dag}Scaling parameters exclude the input embedding parameters.
  The ratio $r$ is nanochat's \texttt{--target-param-data-ratio}, defined as
  $r = \text{training tokens} / \text{scaling params}$.
  All models are trained with FP8 on $8\times$H100 with DDP.
  }
  \label{tab:training-config}
\end{table*}

\subsection{Annotation of Content Errors.} \label{sec:llm-annotate}

Figure \ref{fig:factual-error-prompt} and \ref{fig:reasoning-annotation-prompt} shows the annotation prompts by \texttt{Gemini-3-flash-preview} on factual errors, reasoning steps, and errors.

\begin{figure}[h]

\begin{promptbox}[Factual-error annotation prompt]
factual_errors (structured output)

Identify CLEARLY WRONG verifiable factual claims in the document.
A "factual error" means a verifiable statement (date, number, named
entity, event, attribution) that you are confident is incorrect based
on general knowledge.

STRICT RULES:
  1. Only list errors you are CONFIDENT are wrong. When uncertain,
     do not list. False positives are worse than false negatives.
  2. Only verifiable factual claims count:
     - dates, years, numbers, measurements
     - named entities (people, places, organizations)
     - historical or scientific events
     - attributions (who said/wrote/invented X)
  3. Do NOT list:
     - opinions, values, aesthetic judgments
     - grammatical or stylistic mistakes
     - outdated-but-once-true statements, unless clearly wrong today
     - vague generalizations without specific claims
     - invalid reasoning from correct facts; that is reasoning_errors
  4. For each error, QUOTE the exact text from the document
     (max ~30 words per quote), and briefly explain why it is wrong.
  5. If there are no clear errors, return an empty list.

OUTPUT (JSON field):
  "factual_errors": [
    {
      "quote": "<exact text from the document, <=30 words>",
      "why_wrong": "<one-sentence reason, what the correct fact is>"
    },
    ...
  ]

If the document has no clear errors, return:
  "factual_errors": []
\end{promptbox}
\vspace{-1.0em}
\caption{Prompt for factual errors annotation.}
\label{fig:factual-error-prompt}
\vspace{-1.0em}
\end{figure}

\begin{figure*}[t]
\centering
\begin{minipage}[t]{0.49\textwidth}
\begin{promptbox}[Reasoning steps]
For each document CHUNK, identify reasoning_steps:
each place where the chunk makes a logical inference
(not just states a fact).

A "reasoning step" is an inference: a move from premises to a
conclusion, a derivation, a calculation, a causal attribution, an
analogy used as argument, or any other logical bridging. Mere
description, narration, or recall of facts is NOT a reasoning step.

Examples of reasoning steps:
  - "Because the temperature is above 100C, the water will boil."
    (premise -> conclusion)
  - "If all primes > 2 are odd, then 97 is odd because 97 is prime."
    (modus ponens on a known property)
  ....

NOT reasoning steps:
  - Pure description: "The Eiffel Tower is 330 m tall."
  - Narrative sequence: "She walked to the store and bought bread."
  - Quotations that merely restate a source.

STRICT RULES:
  1. Count at the granularity of inference moves:
     one premise -> conclusion pair = one step.
     A chained argument with two deductions counts as two steps.
  2. Quote the exact text that performs the inference (<=30 words).
  3. In `explanation`, state what is being inferred in ONE sentence.
  4. If the chunk contains NO reasoning, return an empty list.
\end{promptbox}
\end{minipage}
\hfill
\begin{minipage}[t]{0.49\textwidth}
\begin{promptbox}[Reasoning errors]
For each document CHUNK, identify reasoning_errors:
each place where the reasoning is clearly INVALID.

A "reasoning error" is an invalid logical step. Types that count:
  - non sequitur (conclusion does not follow)
  - affirming the consequent
  - denying the antecedent
  - hasty generalisation
  ....

STRICT RULES:
  1. Only list errors you are CONFIDENT are invalid.
     When uncertain, do not list.
     False positives are worse than false negatives.
  2. A factual error used as a premise counts as a factual error,
     NOT a reasoning error, unless the logical step is also invalid.
  3. Quote the exact invalid step (<=30 words) and explain in ONE
     sentence why it is wrong.
  4. If the chunk contains NO invalid reasoning, return an empty list.

OUTPUT (JSON only):

{
  "reasoning_steps": [
    {"quote": "<<=30 words>", "explanation": "<one sentence>"},
    ...
  ],
  "reasoning_errors": [
    {"quote": "<<=30 words>", "why_wrong": "<one sentence>"},
    ...
  ]
}

No preamble, no explanation outside the JSON, no markdown.
\end{promptbox}
\end{minipage}

\caption{Prompt for reasoning steps and reasoning errors annotation.}
\label{fig:reasoning-annotation-prompt}
\end{figure*}

\begin{figure*}[!t]
    \centering
    \includegraphics[width=\linewidth]{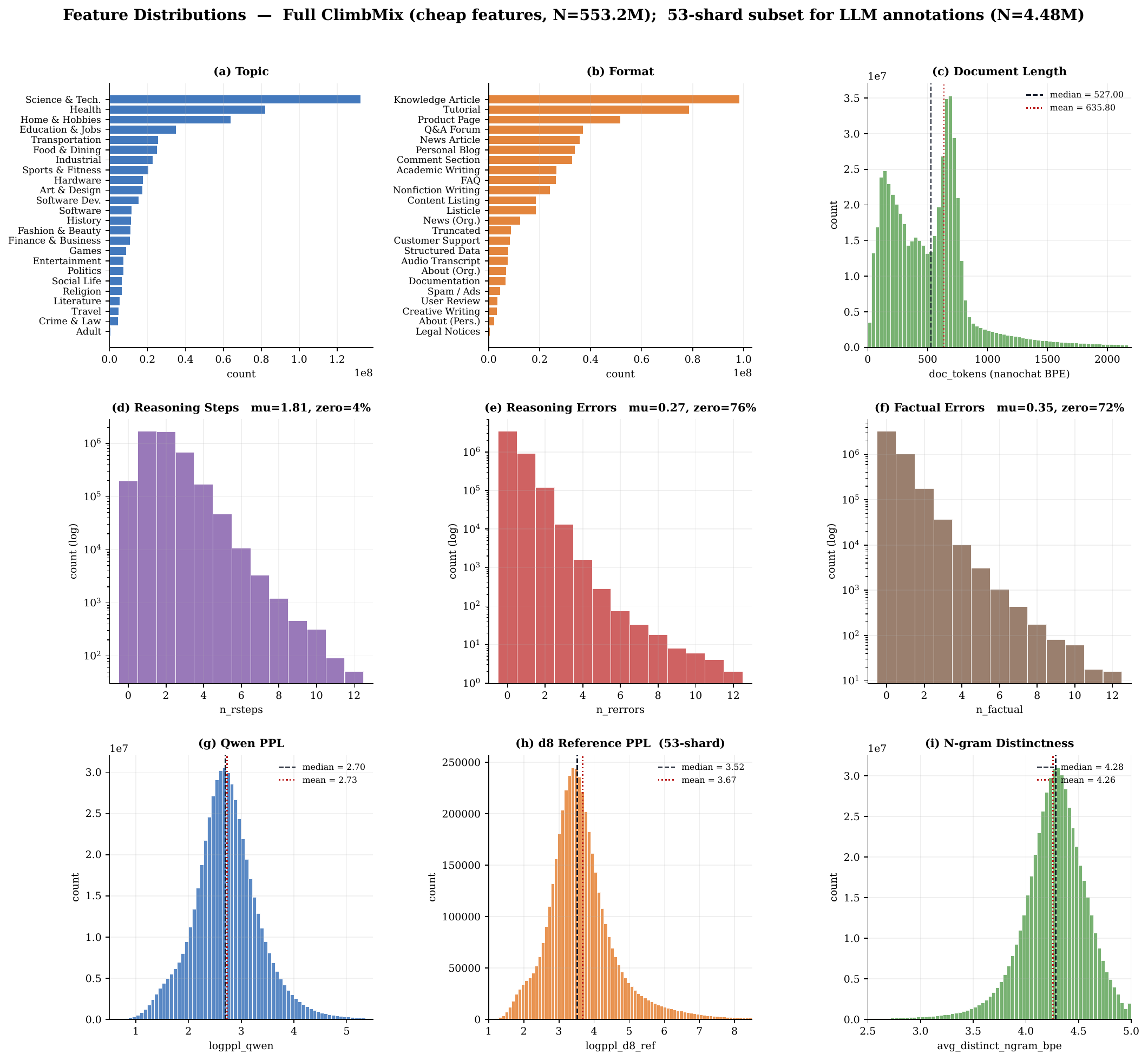}
    \caption{Full ClimbMix data distribution with multiple features. }
    \label{fig:distribution}
\end{figure*}

\begin{figure*}[!t]
    \centering

    \begin{subfigure}{0.48\linewidth}
        \centering
        \includegraphics[width=\linewidth]{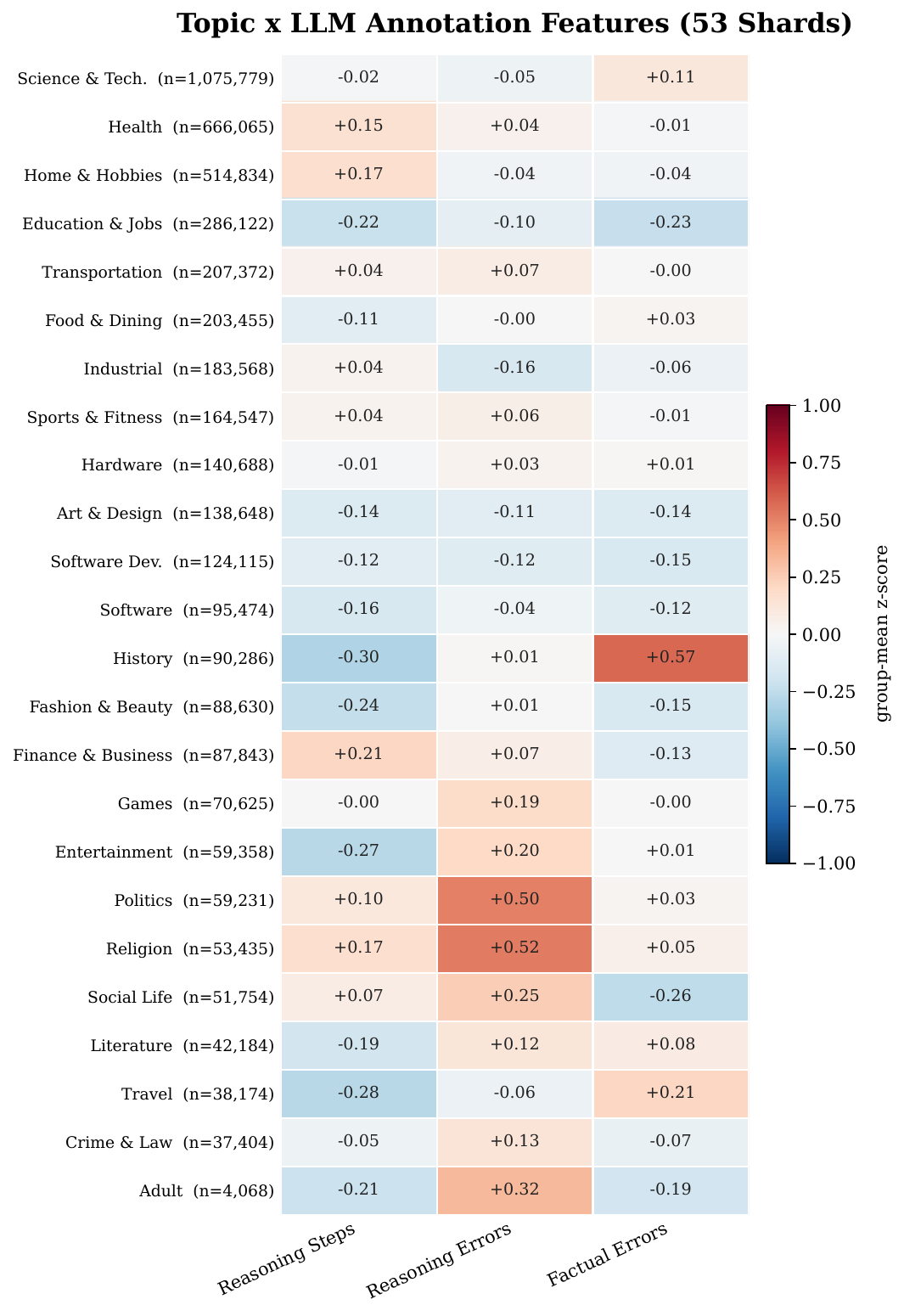}
        \caption{Topic distribution by LLM annotation features.}
        \label{fig:topic-llm-features}
    \end{subfigure}
    \hfill
    \begin{subfigure}{0.48\linewidth}
        \centering
        \includegraphics[width=\linewidth]{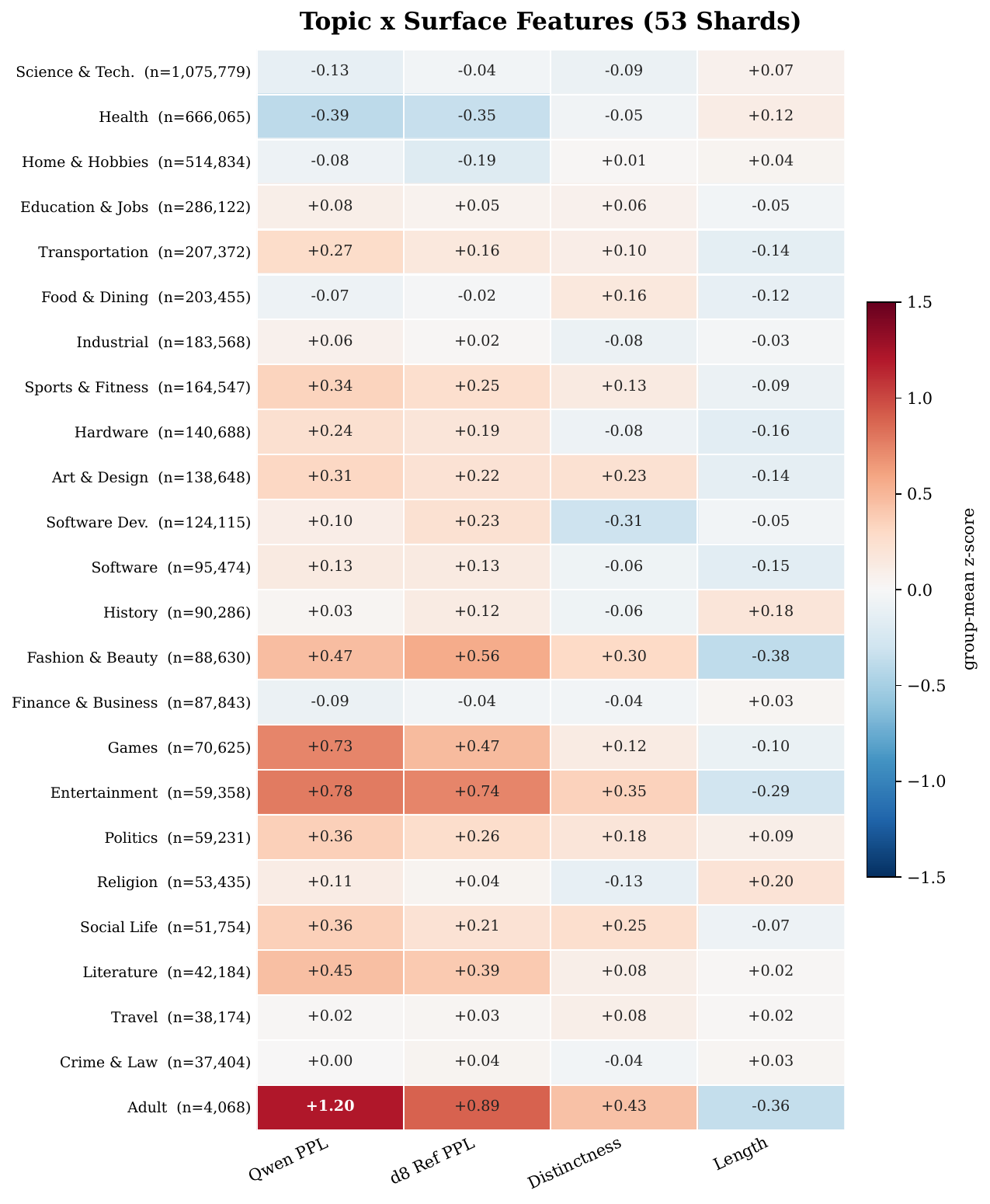}
        \caption{Topic distribution by surface features.}
        \label{fig:topic-surface-features}
    \end{subfigure}

    \caption{Feature correlation in the 53-shard ClimbMix pool.}
    \label{fig:topic-feature-interactions}
\end{figure*}

 \begin{figure*}[t]
  \begin{ideabox}[title={\textbf{Recipe \#1} (\textsc{AutoData (val-bpb)}; GPT-5.5 Step 93,
  $+5.6\sigma$ on \textsc{val-bpb})}]
  \textbf{[Code]}
  \begin{lstlisting}[style=idealisting]
  def select_docs(budget=BUDGET, seed=42):
      rng = np.random.default_rng(seed)
      n_backbone = budget // 3
      n_fill = budget - n_backbone
      arm = min(3, (N - n_backbone) // n_fill)
      draw = rng.choice(N, size=n_backbone + n_fill * arm, replace=False)
      backbone, cand = draw[:n_backbone], draw[n_backbone:]

      # --- Per-candidate quality score ---------------------------------
      div  = (_tanh(div_avg[cand])         + _tanh(div5[cand]))    * 0.5
      ppl  =  _tanh(logppl[cand],     prefer_low=True)
      lenS =  _central_tanh(np.log1p(doc_tokens[cand]))
      ratS =  _central_tanh(doc_chars[cand] / np.maximum(doc_tokens[cand], 1))
      score = div + ppl * (div + 1.0) * 0.5 + lenS + ratS
  
      # --- Shrinkage centering on (topic, format) cells ---------------
      inv     = _category_inverse(topic_id[cand], format_id[cand])
      counts  = np.bincount(inv)
      means   = np.bincount(inv, weights=score) / np.maximum(counts, 1)
      shrink  = counts / (counts + max(np.median(counts), 1.0))
      # --- Gumbel slate-argmax tournament -----------------------------
      score += rng.gumbel(size=score.shape)
      slate  = score.reshape(n_fill, arm)
      winners = cand[np.arange(n_fill) * arm + np.argmax(slate, axis=1)]

      return np.sort(np.concatenate([backbone, winners]))
  \end{lstlisting}

  \end{ideabox}
  \caption{Recipe \#1 --- \textsc{AutoData (val-bpb)} best step}
  \label{fig:recipe2-ideabox}
  \end{figure*}

\begin{figure*}[t]
  \begin{ideabox}[title={\textbf{Recipe \#2} (\textsc{AutoData (CORE)}; GPT-5.5, step 105, $+7.3\sigma$ on \textsc{Core})}]

  \medskip
  \textbf{[Code]}
  \begin{lstlisting}[style=idealisting]
  def select_docs(budget=BUDGET, seed=42):
      rng = np.random.default_rng(seed)
      n_backbone = budget // 2
      n_repair   = budget - n_backbone
      n_cand     = min(N - n_backbone, n_repair * 2)
      draw = rng.choice(N, size=n_backbone + n_cand, replace=False)
      backbone, cand = draw[:n_backbone], draw[n_backbone:]

      # --- Goldilocks tri-anchor scores: -|x - anchor| / (hi - lo) ----
      log_len = np.log1p(doc_tokens[cand])
      ratio   = np.log((doc_chars[cand] + 1) / (doc_tokens[cand] + 1))
      div     = avg_distinct_ngram_bpe[cand]
      ppl     = logppl_qwen[cand]

      lq = _safe_quantiles(log_len, [1/3, 2/3])      # length tertiles
      dq = _safe_quantiles(div,     [1/3, 2/3])      # diversity tertiles
      pq = _safe_quantiles(ppl,     [1/3, 2/3])      # PPL tertiles
      rq = _safe_quantiles(ratio,   [1/3, 1/2, 2/3]) # char/tok with median anchor
  
      s_len = _tri(log_len, anchor=lq[1], lo=lq[0], hi=lq[1])
      s_div = _tri(div,     anchor=dq[1], lo=dq[0], hi=dq[1])
      s_ppl = _tri(ppl,     anchor=pq[0], lo=pq[0], hi=pq[1])  # prefer low-PPL
      s_rat = _tri(ratio,   anchor=rq[1], lo=rq[0], hi=rq[2])
      indiv = _zscore((s_len + s_div + s_ppl + s_rat) / 4.0)
  
      # --- Source-order neighborhood prior ----------------------------
      n_blocks   = max(1, int(np.sqrt(cand.size)))
      block      = np.minimum((cand * n_blocks) // N, n_blocks - 1)
      counts     = np.bincount(block, minlength=n_blocks)
      sums       = np.bincount(block, weights=indiv, minlength=n_blocks)
      block_mean = sums / np.maximum(counts, 1)
      block_prior = _zscore(block_mean)[block]   # rewards consistently clean source ranges
  
      # --- Gumbel-perturbed top-n selection ---------------------------
      total = indiv + block_prior + rng.gumbel(size=cand.size)
      repair = cand[np.argpartition(total, -n_repair)[-n_repair:]]
      return np.sort(np.concatenate([backbone, repair]))
  \end{lstlisting}
  
  \end{ideabox}
  \caption{Recipe \#2 --- \textsc{AutoData (CORE)} best step}
  \label{fig:recipe3-ideabox}
  \end{figure*}

\begin{figure*}[t]
  \begin{ideabox}[title={\textbf{Recipe \#3} (\textsc{AutoData (val-bpb)}, Claude-Opus-4.7, step 93, $+4.8\sigma$ on \textsc{val-bpb})}]

  \medskip
  \textbf{[Code]}
  \begin{lstlisting}[style=idealisting]
  def select_docs(budget=BUDGET, seed=42):
      rng = np.random.default_rng(seed)
      logppl = np.asarray(logppl_qwen).copy()
      logppl[~np.isfinite(logppl)] = float(np.nanmedian(logppl))

      # --- Spam-corner + very-short soft penalty masks ----------------
      spam_mask  = (distinct_5gram_bpe <= np.quantile(distinct_5gram_bpe, 0.10)) \
                 & (logppl             <= np.quantile(logppl,             0.10))
      short_mask =  doc_tokens         <= np.quantile(doc_tokens,         0.05)
  
      # --- (topic, format, length-bin) joint cell key, 4 length bins --
      n_bins  = 4
      edges   = np.quantile(doc_tokens, np.linspace(0,1,n_bins+1)[1:-1])
      len_bin = np.searchsorted(edges, doc_tokens, side="right")
      cell    = (topic_id * (format_id.max()+1) + format_id) * n_bins + len_bin
  
      # --- Gumbel keys with penalties ---------------------------------
      keys = rng.gumbel(size=N) - 1.5 * spam_mask - 0.5 * short_mask
  
      # --- Proportional per-cell quota (largest-remainder) -----------
      order = np.argsort(cell, kind="stable")
      uniq_cells, starts, counts = np.unique(cell[order], return_index=True, return_counts=True)
      raw   = counts * (budget / N)
      quota = np.floor(raw).astype(np.int64)
      extra = budget - quota.sum()
      if extra > 0:                       # distribute leftover via largest-remainder
          rem_top = np.argpartition(-(raw - quota), extra-1)[:extra]
          quota[rem_top] += 1
      quota = np.minimum(quota, counts)   # never overshoot a cell

      # --- Within each cell, take top-q docs by penalised key ---------
      out = np.empty(quota.sum(), dtype=np.int64); w = 0
      for s, c, q in zip(starts, counts, quota):
          if q <= 0: continue
          local = (np.arange(c) if q >= c
                   else np.argpartition(-keys[order[s:s+c]], q-1)[:q])
          out[w:w+q] = order[s + local]; w += q
  
      return np.sort(np.unique(out))
  \end{lstlisting}
  
  \end{ideabox}
  \caption{Recipe \#3 --- \textsc{AutoData (val-bpb)} Opus-4.7 best step}
  \label{fig:recipe4-ideabox}
  \end{figure*}

 \begin{figure*}[t]
  \begin{ideabox}[title={\textbf{Recipe \#4} (\textsc{AutoData (val-bpb)}; Gemini-3-Pro-Preview, step 112, $+4.6\sigma$ on \textsc{val-bpb})}]

  \medskip
  \textbf{[Code]}
  \begin{lstlisting}[style=idealisting]
  def select_docs(budget=BUDGET, seed=42):
      rng = np.random.default_rng(seed)

      # --- 85% robust uniform backbone --------------------------------
      n_backbone = int(budget * 0.85)
      n_repair   = budget - n_backbone
      backbone   = rng.choice(N, size=n_backbone, replace=False)
      pool       = np.setdiff1d(np.arange(N), backbone, assume_unique=False)

      # --- 4x candidate oversample for the 15% quality repair ---------
      cand = np.sort(rng.choice(pool, size=min(n_repair*4, pool.size), replace=False))
  
      # --- 10 global length deciles via random subsample ---------------
      sub      = rng.choice(N, size=1_000_000, replace=False)
      len_bins = np.quantile(doc_tokens[sub], np.linspace(0,1,11))
      len_bins[0], len_bins[-1] = -1, np.inf
      cand_bin = np.clip(np.digitize(doc_tokens[cand], len_bins) - 1, 0, 9)
  
      # --- Per-bin Spearman-Mahalanobis distance to ideal percentile --
      target = np.array([1/3, 0.85, 0.85, 0.50])    # PPL, AvgDiv, 5gDiv, Char/Tok
      repair = []
      for b in range(10):
          m = cand_bin == b
          b_idx = cand[m]
          q = min(int(np.ceil(n_repair * m.sum() / cand.size)), b_idx.size)
          if q == 0: continue
  
          feats = np.column_stack([
              np.where(np.isnan(logppl_qwen[b_idx]),         np.inf, logppl_qwen[b_idx]),
              np.where(np.isnan(avg_distinct_ngram_bpe[b_idx]), -np.inf, avg_distinct_ngram_bpe[b_idx]),
              np.where(np.isnan(distinct_5gram_bpe[b_idx]),    -np.inf, distinct_5gram_bpe[b_idx]),
              doc_chars[b_idx] / np.maximum(doc_tokens[b_idx], 1),
          ])
          # Map each feature to in-bin uniform percentiles
          U = np.argsort(np.argsort(feats, axis=0), axis=0) / max(len(b_idx)-1, 1)
          # Spearman-Mahalanobis distance to ideal target
          invC  = np.linalg.pinv(np.cov(U, rowvar=False))
          delta = U - target
          dist  = np.einsum("ij,jk,ik->i", delta, invC, delta)
          # Reject malformed (NaN/inf-flagged) docs
          dist[np.isinf(feats[:,:3]).any(axis=1)] = np.inf
          repair.append(b_idx[np.argsort(dist)[:q]])
  
      repair = np.concatenate(repair) if repair else np.empty(0, dtype=np.int64)
      out    = np.concatenate([backbone[: budget - repair.size], repair])
      return np.sort(out.astype(np.int64))
  \end{lstlisting}

  \end{ideabox}
  \caption{Recipe \#4 --- \textsc{AutoData (val-bpb)} Gemini-3-Pro-Preview best step}
  \label{fig:recipe5-ideabox}
  \end{figure*}

\end{document}